\documentclass[letterpaper, 10 pt, conference]{ieeeconf}  

\IEEEoverridecommandlockouts                              

\usepackage{graphics} 
\usepackage{epsfig} 
\usepackage{mathptmx} 
\usepackage{times} 
\usepackage{amsmath} 
\usepackage{amssymb}  
\usepackage[at]{easylist}
\usepackage{setspace}
\usepackage{soul}
\usepackage{color}
\usepackage[binary-units=true]{siunitx}
\usepackage{xcolor}
\usepackage{mathtools}
\usepackage{float}
\usepackage[ruled,vlined,linesnumbered,noresetcount]{algorithm2e}
\usepackage[pagebackref=true,breaklinks=true,colorlinks,bookmarks=false]{hyperref}

\title{\LARGE \bf Hybrid Differential Dynamic Programming for Planar Manipulation Primitives\vspace{-2ex}}

\author{
\authorblockN{Neel Doshi$^{1,2}$ Francois R. Hogan$^2$, Alberto Rodriguez$^2$}
\authorblockA{{\tt\small <nddoshi, fhogan, albertor>@mit.edu}}
\thanks{\hspace{-0.3cm}$^1$Intelligence Community Postdoctoral Research Fellowship Program, Massachusetts Institute of Technology, Cambridge, MA \newline
$^2$Dept. of Mechanical Engineering, Massachusetts Institute of Technology}\vspace{-1cm}}

\def\*#1{\mathbf{#1}}
\def\?#1{\mathbb{#1}}

\newcommand{\myparagraph}[1]{\vspace{0.05in}\noindent\textbf{#1}}

\begin{document}

\maketitle

\thispagestyle{empty}
\pagestyle{empty}

\begin{abstract}
We present a hybrid differential dynamic programming (DDP) algorithm for closed-loop execution of manipulation primitives with frictional contact switches. Planning and control of these primitives is challenging as they are hybrid, under-actuated, and stochastic. 
We address this by developing hybrid DDP both to plan finite horizon trajectories with a few contact switches and to create linear stabilizing controllers.
We evaluate the performance and computational cost of our framework in ablations studies for two primitives: planar pushing and planar pivoting. We find that generating pose-to-pose closed-loop trajectories from most configurations requires only a couple (one to two) hybrid switches and can be done in reasonable time (one to five seconds). 
We further demonstrate that our controller stabilizes these hybrid trajectories on a real pushing system. 
A video describing our work can be found at \url{https://youtu.be/YGSe4cUfq6Q}. 
\end{abstract}

\section{Introduction}

Complex manipulation tasks can often be decomposed into a sequence of simpler behaviors. For example, picking a credit-card off a table may consist of a pull to cantilever the card followed by a grasp to acquire the card. In part motivated by this observation, researchers have studied manipulation behavior segmented into \textit{manipulation primitives} such as grasping, pulling, pushing, or pivoting. 

These primitives are often used to facilitate planning and control; however, defining  these primitives, planning within a primitive, and scheduling primitives are all current areas of research. One approach is to use narrowly defined primitives that are simpler to plan and control at the expense of needing more of them; for example, when every contact mode/type is a primitive \cite{woodruff2017planning}. On the other hand, complex, more expressive primitives often incur a higher computational cost and can be challenging to realize on a physical system. An exteme case is when all possible mode sequences are considered in the development of a single behavior \cite{Posa14, manchester2019contact}. 

This work proposes a fast planning and control framework that supports a small number of hybrid switches for primitives of moderate complexity with underactuated frictional dynamics. Switching contact formations within a primitive increases its expressiveness, which can reduce the number of primitives needed and, consequently, ease their scheduling.
    
\myparagraph{Contributions}
We develop an algorithm for executing manipulation primitives with frictional contact switches. Our approach extends input-constrained differential dynamic programming (DDP) to handle hybrid dynamics. We plan finite-horizon trajectories while considering a small number of contact switches (up to four) within a reasonable amount of time (one to five seconds). We also present a numerical study on the convergence properties and computational requirements of our algorithm for two manipulation primitives: planar pushing and planar pivoting. Our experiments find that
\begin{itemize}
\item The ability to select and switch contact locations is key to the success of a primitive.
\item Only one to two contact location switches are needed to converge from most initial configurations.
\end{itemize}
Finally, we show that our framework can plan and control hybrid trajectories on a real planar pushing system. 

\myparagraph{Paper Structure}
We begin by reviewing the basics of DDP, extensions to handle constraints, and our hybrid algorithm in Sec.~\ref{sec:planning_and_control}. We then derive the motion models for planar pushing and pivoting in Sec.~\ref{sec:example_primitives}. Section~\ref{sec:sim_results} describes simulation studies that evaluate the success rate and computation time of the algorithm for these two primitives. 
We experimentally validate the algorithm for the planar pushing primitive in Sec.~\ref{sec:exp_results}, and finally, we summarize the results, limitations, and directions of future work in Sec.~\ref{sec:discussion}. 
    
\section{Related Work}
\label{sec:lit_review}

In this section we discuss some related research on manipulation primitives and DDP.

\myparagraph{Manipulation Primitives}
There is a long history in robotic manipulation of developing the mechanics of and planning algorithms for primitives. Mason~\cite{mason1986mechanics} introduced the mechanics of planar pushing, which have since been studied by a number of researchers~\cite{lynch1996stable, hogan2016wafr,zhou2017pushing}. This line of work has been extended to many other primitives, including prehensile-pushing~\cite{chavan2018hand}, tumbling~\cite{sawasaki1991tumbling}, pivoting~\cite{holladay2015general, karayiannidis2016adaptive, hou2018fast}, scooping~\cite{trinkle1988investigation, trinkle1992stability}, tilting~\cite{erdmann1988exploration}, and dynamic in-hand sliding~\cite{shi2017dynamic}. 

Researchers have also focused on sequencing primitives to achieve complex manipulations \cite{trinkle1991framework, terasaki1998motion, Barry2013, suarez2016framework, hogan2020tactile}. For example, Toussaint et al.~\cite{toussaint2018differentiable} use a few kinematic primitives to realize diverse set of behaviors; however, this approach is only verified in simulation. Woodruff et al.~\cite{woodruff2017planning} treat each contact formation as a different primitive. They then execute closed-loop dynamic motions with a fixed primitive-schedule using a planar manipulator in a low-gravity environment. Our framework balances these approaches and is similar to that of Hou et al.~\cite{hou2018fast}, who develop a planner for two moderate-complexity primitives and demonstrate pose-to-pose re-orientation on a physical system.

\myparagraph{Differential Dynamic Programming}
DDP is an iterative, indirect trajectory optimization method that leverages the  structure in Bellman's equation to achieve local optimality. Originally developed by Jacobson and Mayne~\cite{jacobson1970differential} for unconstrained systems, it has since been extended to systems with box input constraints~\cite{tassa2014control}, linear input constraints~\cite{murray1979constrained}, and non-linear constraints on input and states~\cite{xie2017differential}. 

Relevant to this work, Tassa et al.~\cite{Tassa12} and Mordatch et al.~\cite{Mordatch12} use DDP with smoothed contact models to plan and stabilize trajectories for legged robots. Yamaguchi and Atkeson~\cite{Yamaguchi2016} apply DDP to the problem of planning for graph-dynamical systems, and they use a sample based approach to determine the mode sequence. Moreover, Pajarinen et al.~\cite{pajarinen2017hybrid} consider DDP for planar pushing, and they optimize over a continuous mixture of discrete actions that is forced back into fully discrete actions at convergence. Our work improves on these approaches by considering exact rigid-body frictional contacts, determining the mode schedule using DDP, and retaining the anytime property of the algorithm.

\section{Hybrid Planning and Control}
\label{sec:planning_and_control}

We first review the basics of DDP (Sec.~\ref{sec:ddp_basics}) and its extension to input-constrained systems (Sec.~\ref{sec:ddp_input_cnst}). We then describe our hybrid DDP algorithm in Sec.~\ref{sec:hybrid_ddp}. 

\subsection{DDP Pereliminaries}
\label{sec:ddp_basics}

Consider a discrete-time dynamical system of the form 
\begin{align}
\label{eq:ddp_dyn}
\*{x}_{k+1} &= \*{f}(\*{x}_k, \*{u}_k)
\end{align}
where $\*{f}$ is a smooth function that maps the system's state ($\*{x} \in \?{R}^n$) and control input ($\*{u} \in \?{R}^m$) to the next state. The goal is to find an input trajectory $\*{U} \coloneqq  \{\*{u}_0, \*{u}_1, \hdots, \*{u}_{N-1} \}$ that minimizes an additive cost function, 
\begin{equation}
\label{eq:total_cost}
J(\*{x}_0, \*{U}) = l_f(\*{x}_N) + \sum_{k=0}^{N-1} l(\*{x}_k, \*{u}_k). 
\end{equation} 
Here $k$ is the time-step, $l$ is the running cost, $l_f$ is the final cost, $N$ is the time-horizon, $\*{x}_0$ is the initial state, and $\*{x}_1 \hdots \*{x}_N$ are determined by integrating \eqref{eq:ddp_dyn} forward in time. We can define the \textit{optimal cost-to-go} at the $k$-th time-step using Bellman's equation \cite{bellman1966dynamic},
\begin{align}
\label{eq:cost_to_go}
V_k(\*{x}_k) = \min_{\*{u}_k} [l(\*{x}_k, \*{u}_k) + V_{k+1}(\*{f}(\*{x}_k, \*{u}_k))], 
\end{align} 
with the terminal condition $V_N(\*{x}_N) = l_f(\*{x}_N)$. 

To handle the non-linearity in \eqref{eq:cost_to_go}, DDP iteratively optimizes a quadratic approximation near an initial trajectory. The algorithm starts with a forward pass that integrates \eqref{eq:ddp_dyn} from an initial state $\*{x}_0$ using a current guess of the input trajectory $\*{U}$. This is followed by a backward pass that computes a local solution to \eqref{eq:cost_to_go} using a quadratic Taylor expansion to iterate on the value of $\*{U}$. This sequence of forward and backward passes is repeated until convergence. 

The Taylor expansion of the argument of \eqref{eq:cost_to_go} about a nominal $(\*{x}, \*{u})$ pair is given by
\begin{align}
\label{eq:local_ctg}
    Q(\*{\delta x}, \*{\delta u}) &= l(\*{x} + \*{\delta x}, \*{u} + \*{\delta u})  -  l(\*{x}, \*{u}) \nonumber \\ 
    & + V_{k+1}(\*{f}(\*{x} + \*{\delta x}, \*{u} + \*{\delta u})) - V_{k+1}(\*{f}(\*{x}, \*{u})).
\end{align}
The quadratic approximation of $Q$ can be written as: 
\begin{align}
\label{eq:quad_approx}
Q(\*{\delta x}_k, \*{\delta u}_k) \approx \frac{1}{2} 
\begin{bmatrix} 1 \\ \*{\delta x}_k \\  \*{\delta u}_k \end{bmatrix}^T
\begin{bmatrix} 0 & \*{q}_\*{x}^T & \*{q}_\*{u}^T \\ 
\*{q}_\*{x} & \*{Q}_\*{xx} & \*{Q}_\*{xu} \\   
\*{q}_\*{u} & \*{Q}_\*{xu}^T & \*{Q}_\*{uu} \\  
\end{bmatrix}
\begin{bmatrix} 1 \\ \*{\delta x}_k \\  \*{\delta u}_k \end{bmatrix}, 
\end{align}
where the block matrices are functions of $V_{k+1}$, $l$, $\*{f}$, and their first and second derivatives \cite{tassa2014control}. The control modification is obtained by minimizing \eqref{eq:quad_approx} with respect to $\*{\delta u}$ for some state perturbation $\*{\delta x}$:
\begin{align}
\label{eq:control_update}
\*{\delta u^*} = -\*{Q}_\*{uu}^{-1}(\*{q}_\*{u} +  \*{Q}_\*{xu}^T\*{\delta x} ) = \*{k} + \*{K} \*{\delta x}, 
\end{align}
where $\*{k} =  -\*{Q}_\*{uu}^{-1}\*{q}_\*{u}$ is the feed-forward control and $\*{K} = -\*{Q}_\*{uu}^{-1}\*{Q}_\*{xu}^T$ is the feedback gain. Substituting this for $\*{\delta u}$ in \eqref{eq:quad_approx} gives a quadratic model for $V$
\begin{align}
\label{eq:value_update}
\Delta V &= \frac{1}{2} \*{k}^T \*{Q}_\*{uu} \*{k} \nonumber \\
\*{V}_\*{x} &= \*{q}_\*{x} - \*{K}^T  \*{Q}_\*{uu} \*{k} \nonumber \\
\*{V}_\*{xx} &= \*{Q}_\*{xx} - \*{K}^T  \*{Q}_\*{uu} \*{K}. 
\end{align}
The backward pass initializes the quadratic approximation of $V$ with $l_f(\*{x}_N)$ and its derivatives, and then recursively computes \eqref{eq:control_update} and propagates the value approximation \eqref{eq:value_update}. 

\begin{algorithm}[t]
\SetAlgoLined
initialize $\leftarrow \*{x}_0,\*{U}_0$ \\
\While{not converged} {
$V_N \leftarrow l_f(\*{x}_N)$ \\
\For{$k=N-1$ \KwTo $0$ }{
    $Q_k (\*{\delta x}_k, \*{\delta u}_k) \leftarrow$ \eqref{eq:quad_approx} \\
    $\*{k} \leftarrow$ solve QP \eqref{eq:bp_qp};  $\*{K} \leftarrow$ see \cite{murray1979constrained} \\
    $\*{\delta u}_k \leftarrow \*{k} + \*{K} \*{\delta x_k}$ \\
    Propagate value $\leftarrow$ \eqref{eq:value_update} \\
}
$\*{\hat{x}}_0 \leftarrow \*{x}_0$ \\  
\For{$k=0$ \KwTo $N-1$}{
    $\*{\hat{u}}_k \leftarrow$ ProjectFeasible\big($\*{u}_k + \*{k} + \*{K}(\*{\hat{x}}_k - \*{x}_k)$\big)\\
    $\*{\hat{x}}_{k+1} = \*{f}(\*{\hat{x}}_k, \*{\hat{u}}_k)$
}
$\*{X} \leftarrow \*{\hat{X}}$,  $\*{U} \leftarrow \*{\hat{U}}$
}
\caption{Input constrained DDP}
\label{alg:cnst_ddp}
\vspace{-0.5cm}
\end{algorithm}

The algorithm then integrates \eqref{eq:ddp_dyn} to compute a new trajectory, completing one iteration. The control during this forward pass is set to $\*{u} + \*{\delta u^*}$ with $\delta\*{x_k}$ taken as the difference between $\*{x_k}$ across consecutive iterations. Note that $\*{Q_{uu}}$ is regularized to ensure that $\*{Q_{uu}^{-1}}$ exists and a line-search over $\*{k}$ keeps $\*{\delta u}$ and $\*{\delta x}$ small to ensure cost-reduction. These steps enable convergence from an arbitrary initialization \cite{tassa2014control}.

\subsection{Input Constrained DDP}
\label{sec:ddp_input_cnst}
Now consider a system where the control inputs are linearly constrained by inequality (or equality) constraints:
\begin{align}
\label{eq:constrained_control}
\*{A}(\*{x}_k) \*{u}_k \geq \*{b}(\*{x}_k). 
\end{align}
Here $\*{A}$ and $\*{b}$ are potentially nonlinear functions of the state. This class of constraints can represent both planar friction and force-balance constraints for a fixed contact mode. The DDP algorithm is modified in two ways for these constraints~\cite{murray1979constrained}. First, \eqref{eq:control_update} is replaced by a constrained quadratic program (QP) evaluated at the nominal $\*{x}$:
\begin{align}
\label{eq:bp_qp}
\min_{\*{\delta u}} \quad &Q(
\*{0}, \*{\delta u}) \nonumber \\ 
\text{s.t.}  \quad \quad &\*{A}(\*{x}) (\*{u} + \*{\delta u}) \geq \*{b}(\*{x})
\end{align}
The solution to this QP gives the value of the feed-forward control $\*{k}$ satisfying the input constraints. We then consider the state variation $\*{\delta x}$ when solving for the feedback gain $\*{K}$. Details on this are given by Murray and Yakowitz \cite{murray1979constrained}. 

Second, even though $\*{k}$ satisfies the input constraints, the new control computed using during the forward pass (Line 12, Alg.~\ref{alg:cnst_ddp}) can violate feasibility. Consequently, it must be projected onto the constraint set. When $\*{A}$ and $\*{b}$ have a simple geometric representation (e.g., a box or a cone), we can algebraically project the new control input onto the feasible set. In other cases, we solve another QP detailed by Murray and Yakowitz \cite{murray1979constrained}. The input-constrained DDP algorithm is outlined in Alg.~\ref{alg:cnst_ddp}. 

\subsection{Hybrid DDP}
\label{sec:hybrid_ddp}

We extend input-constrained DDP to systems with hybrid switches. We use DDP as a subroutine to (a) explore and rank all feasible mode sequences and (b) optimize the trajectory and feedback law associated with the best mode sequence. In addition to the initial state and input trajectory, the user can specify the maximum number of hybrid switches ($N_\text{s}$) and the set of hybrid modes ($\mathcal{M}$) of dimension $M$. 

\begin{figure}[t]
    \centering
    \includegraphics[width=0.9\columnwidth]{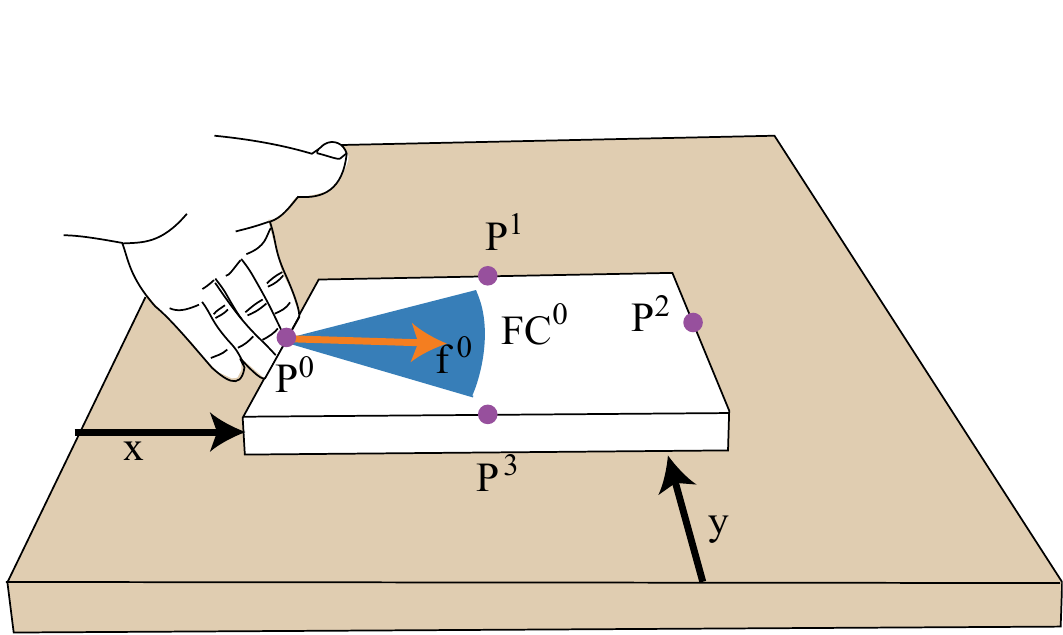}
    \vspace{-0.3cm}
    \caption{Planar pushing with four sticking contacts ($P_0$, $P_1$, $P_2$, $P_3$) at the center of each side. Only one contact can be active at a time, and the active contact force ($\*{f}^i$, $i \in  \{0, 1, 2, 3\}$) must lie within its friction cone ($\text{FC}^i$). }
    \label{fig:planar_pushing}
    \vspace{-0.35cm}
\end{figure}

Our algorithm first builds a depth $N_\text{s} + 1$ tree of length-$N_\text{s}$ possible sequences of contact states. The number of leafs in the tree is upper-bounded by $N_\text{s}^M$. Each leaf is a trajectory associated with a fixed contact mode sequence with the $N_\text{s}$ switch locations distributed evenly along the planning horizon ($N$).  We use input-constrained DDP with a small iteration limit to optimize each leaf in the tree and approximate its cost. We then select the leaf with the lowest cost and fix the mode sequence to that of the selected leaf. Finally, we optimize the trajectory and controller associated with the best leaf using input-constrained DDP (Alg~\ref{alg:cnst_ddp}).

For computational efficiency, we initialize DDP with inputs that result in static equilibrium and prune the tree during exploration by eliminating trajectories that cannot satisfy static equilibrium after a contact switch. The hyper-parameters of our algorithm (explored in Sec.~\ref{sec:sim_results}) are the maximum number of hybrid switches ($N_\text{s}$), the set of hybrid modes ($\mathcal{M}$), the planning horizon ($N$), and the maximum number of DDP iterations during tree generation ($N_\text{i})$. 

In summary, our algorithm can be thought of as an exhaustive tree-search over mode sequences with pruning based on static equilibrium.

\section{Manipulation Primitives}
\label{sec:example_primitives}
Here we derive the equations-of-motion (EOM) for the primitives used in this work: quasi-static planar pushing (Sec.~\ref{sec:pushing}) and dynamic planar pivoting (Sec.~\ref{sec:pivoting}). Our choice of primitives is meant to illustrate that our approach handles manipulation in the horizontal and gravitational planes, as well as, quasi-static and dynamic systems. 
 
\subsection{Quasi-static Planar Pushing}
\label{sec:pushing}

\begin{figure}[t]
    \centering
    \includegraphics[width=0.9\columnwidth]{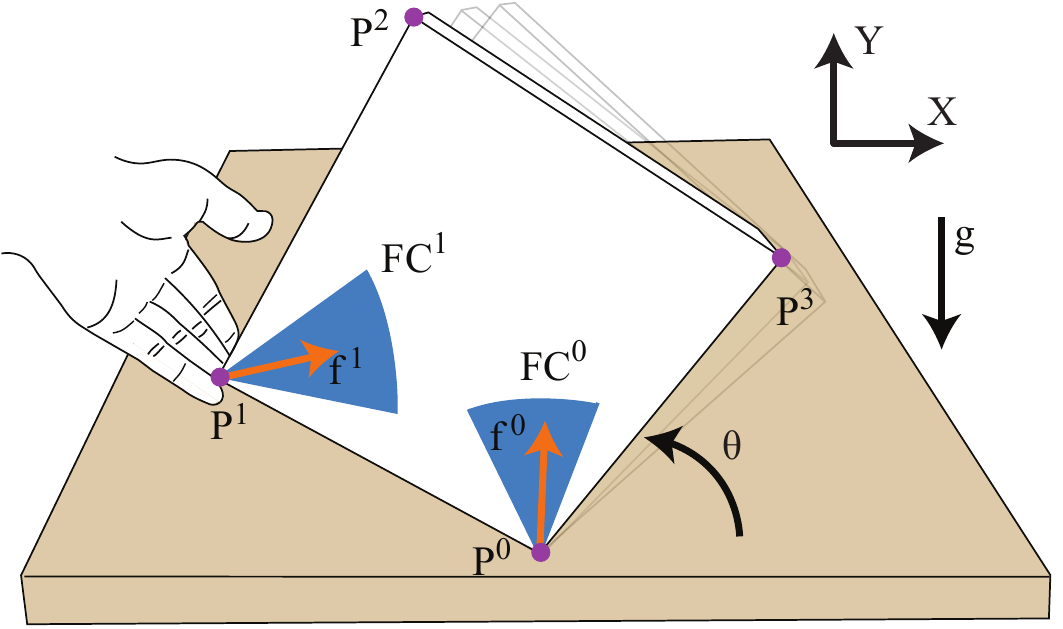}
    \vspace{-0.3cm}
    \caption{Planar pivoting in the gravity plane with the pivot at the lower left corner ($P_0$). We consider three sticking contacts at the other corners ($P_1, P_2, P_3$). Only one contact is active at a time, and the ground-reaction  ($\*{f}^0$) and active contact ($\*{f}^i$, $i \in \{1, 2, 3\}$) forces obey Coulomb friction.}
    \label{fig:planar_pivoting}
    \vspace{-0.35cm}
\end{figure}

We consider quasi-static pushing in a horizontal plane with four potential (only one active at a time) sticking point contacts (Fig.~\ref{fig:planar_pushing}). The object's state is $ \*{x} = [ x, y, \theta ]^T$, where $x$ and $y$ are the position of its center-of-mass (COM) and $\theta$ is its orientation. The discrete-time, quasi-static EOM are 
 \begin{align}
    \label{eq:planar_pushing1}
    \*{x}_{k + 1} = \*{x}_k + \Delta t \*{\dot{x}}_k, 
\end{align}
 where $\*{\dot{x}}_k$ is the object's twist at time-step $k$ and $\Delta t$ is the time step's duration. Using force balance and a ellipsoidal approximation \cite{howe1996practical} of the \textit{limit surface} \cite{goyal1991planar}, we can write
\begin{align}
    \label{eq:planar_pushing2}
    \*{\dot{x}}_k = \*{R}(\theta) \*{L} (\*{J}^{i})^T \*{f}^i_k. 
\end{align}
Here $i \in \{0, 1, 2, 3\}$ is the index of the active contact, $\*{f}^i$ is the active contact force in the body-frame, $\*{J}^i$ is the Jacobian for the active contact, $\*{R}$ is the rotation between body and world frames, and $\*{L}$ is the gradient of the limit surface w.r.t the support wrench. Hogan et al. \cite{hogan2018icra} give further details.  

The $i$-th contact force must lie within its friction cone ($\text{FC}^i$) for sticking contact, and we upper-bound the normal contact force by $N^i_\text{max}$. In a frame whose positive $x$-axis is aligned with the contact normal (i.e., the contact frame),
\begin{align}
    \label{eq:friction_constraint}
    0 \leq f^i_n \leq N_\text{max}^i \nonumber \\
    -\mu f_n^i \leq f^i_t \leq \mu f_n^i. 
\end{align}
Here ($f^i_n, f^i_t$) are the normal and tangential components of $\*{f}^i$ in the contact frame, and $\mu$ is the coefficient of friction.

\subsection{Dynamic Planar Pivoting}
\label{sec:pivoting}

We also consider dynamic pivoting in the gravity plane about a sticking frictional pivot with the ground (Fig.~\ref{fig:planar_pivoting}). The object is rotated about this pivot by sticking contacts at one of the other three corners. Each contact is treated as a point-on-line contact with the line fixed at \SI{45}{\degree} with respect to both sides of that corner. The object's state is $\*{x} = [\theta, \dot{\theta} ]^T$, where $\theta$ is its orientation and $\dot{\theta}$ is its angular velocity. We write the discrete-time dynamics of the system as
\begin{align}
    \*{x}_{k + 1} = \*{x}_k + \Delta t \*{\dot{x}}_k,
    \label{eq:planar_pivot1}
\end{align}
where $\*{\dot{x}}_k = [\dot{\theta}_k, \ddot{\theta}_k]^T$ and $\ddot{\theta}$ is 
\begin{align}
    \label{eq:angular_acceleration}
    \ddot{\theta} = \frac{1}{I} \big(\*{r}^0 \times \*{R}(\theta)^T \*{f}^0 + \*{r}^i \times\*{f}^i\big)
\end{align}
from Newton's angular momentum principle. Here $i \in \{1, 2, 3\}$ is the index of the active contact, $\*{f}^i$ is the active contact force in the body-frame, $\*{f}^0$ is the ground reaction force in the world-frame, $\*{R}^T$ is the rotation between world and body frames, $I$ is the object's mass moment of inertia, and $\*{r}^0$ ($\*{r}^i$) is the vector from the COM to the ground (active) contact. We also constrain the ground reaction force using Newton's linear momentum principle, 
\begin{align}
     \label{eq:momentum_princ}
    \*{\dot{p}}_c = \*{f}^i + \*{R}(\theta)(\*{f}^0 + \*{g}), 
\end{align}
where $\*{g}$ is the gravitational force and $\*{\dot{p}}_c \in \?{R}^2$ is the time-derivative of linear momentum of the COM. Note that $\*{\dot{p}}_c$ can be computed in terms of $\theta,\dot{\theta}$, and $\ddot{\theta}$. Finally, we enforce that all contact forces (active and ground) lie within their friction cones and place an upper bound on the normal contact forces. 

\section{Numerical Studies}
\label{sec:sim_results}

\begin{figure}[t]
\centering
\includegraphics[width=\columnwidth]{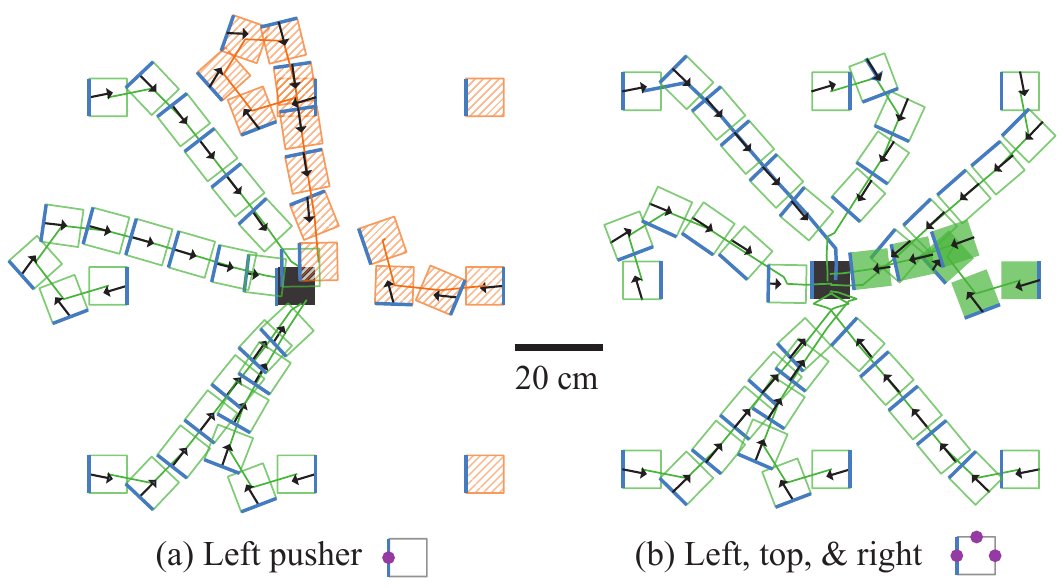}
\vspace{-0.5cm}
\caption{Pushing trajectories from eight initial conditions. The goal is a solid gray square, contact forces are drawn with black arrows, and the left side of object is shown in blue. Successful (unsuccessful) trajectories are depicted in green (patterned orange). The green shaded  trajectory in (b) has a contact switch. }
\label{fig:pp_traj}
\vspace{-0.6cm}
\end{figure}

We use our algorithm to plan pose-to-pose trajectories for both primitives. We present a number of representative trajectories in Sec.~\ref{sec:rep_traj} and conduct ablation studies in Sec.~\ref{sec:ablation}. We use a simple quadratic total cost of the form
\begin{equation}
\label{eq:quad_obj}
 J(\*{x}_0, \*{U}) = \Delta\*{x}_N^T\*{Q}_N\Delta\*{x}_N + \sum_{k=0}^{N-1} \Delta\*{x}_k^T \*{Q} \Delta\*{x}_k + \*{u}^T \*{R} \*{u}, 
\end{equation}
to generate all trajectories. Here $\Delta \*{x}$ is the distance to the goal and $\*{Q}_N$, $\*{Q}$, and $\*{R}$ are positive definite diagonal matrices. 

\subsection{Simulated Trajectory Planning}
\label{sec:rep_traj}

\begin{figure}[t]
\centering
\includegraphics[width=0.8\columnwidth]{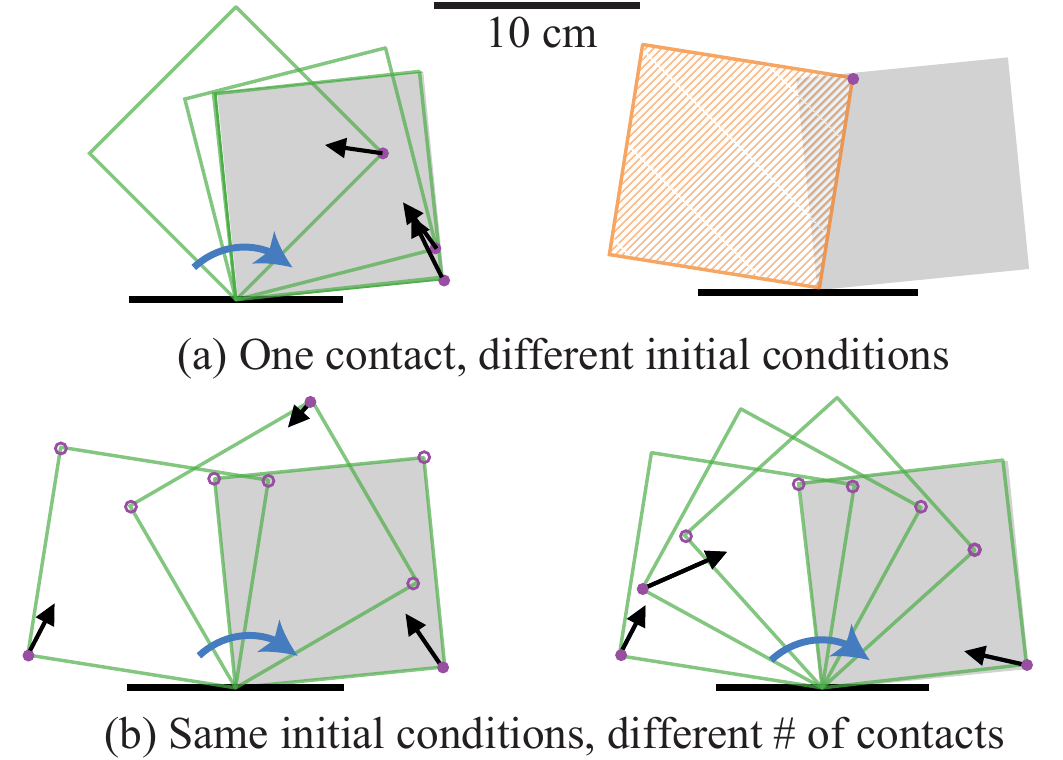}
\vspace{-0.3cm}
\caption{Pivoting trajectories from two representative initial conditions. The goal is the gray square, the direction of rotation is clockwise (blue arrow), the contact forces are drawn with black arrows, and enabled corner-contacts are marked with purple circles with active contacts filled in. Successful (unsuccessful) trajectories are depicted in green (patterned orange).}
\label{fig:pl_traj}
\vspace{-0.3cm}
\end{figure}

Representative planar pushing and pivoting trajectories are shown in Fig.~\ref{fig:pp_traj} and Fig.~\ref{fig:pl_traj}, respectively. 

\begin{figure*}[ht]
\centering
\includegraphics[width=0.9\textwidth]{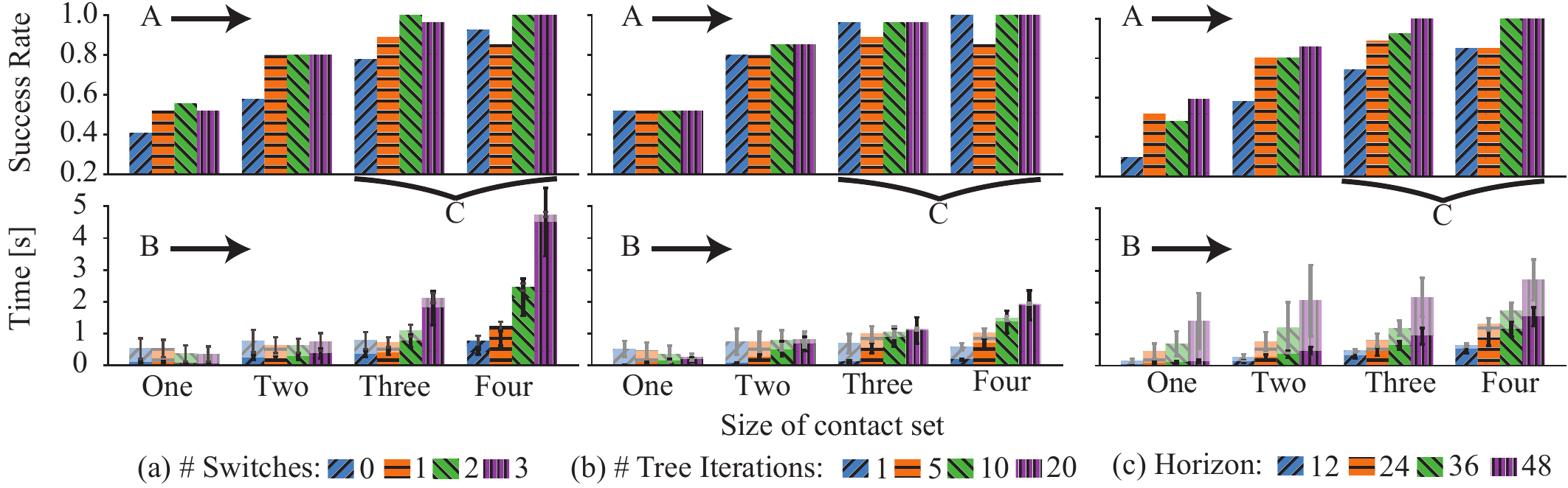}
\vspace{-0.3cm}
\caption{Ablation studies for planar pushing. The top row shows the success rate, and the bottom row shows the planning time for generating the trajectory-tree (solid) and optimizing the best trajectory (transparent). The error bars depict $\pm$ one standard deviation. On the x-axis we show the size of contact set, and the different colors indicate (a) the maximum number of hybrid switches, (b) the number of DDP iterations used when generating the trajectory-tree and (c) the total horizon of the trajectory in discrete steps. The labels A-C highlight important trends that are discussed in the text.}
\label{fig:pp_ablation}
\end{figure*}

\begin{figure*}[ht]
\centering
\includegraphics[width=0.9\textwidth]{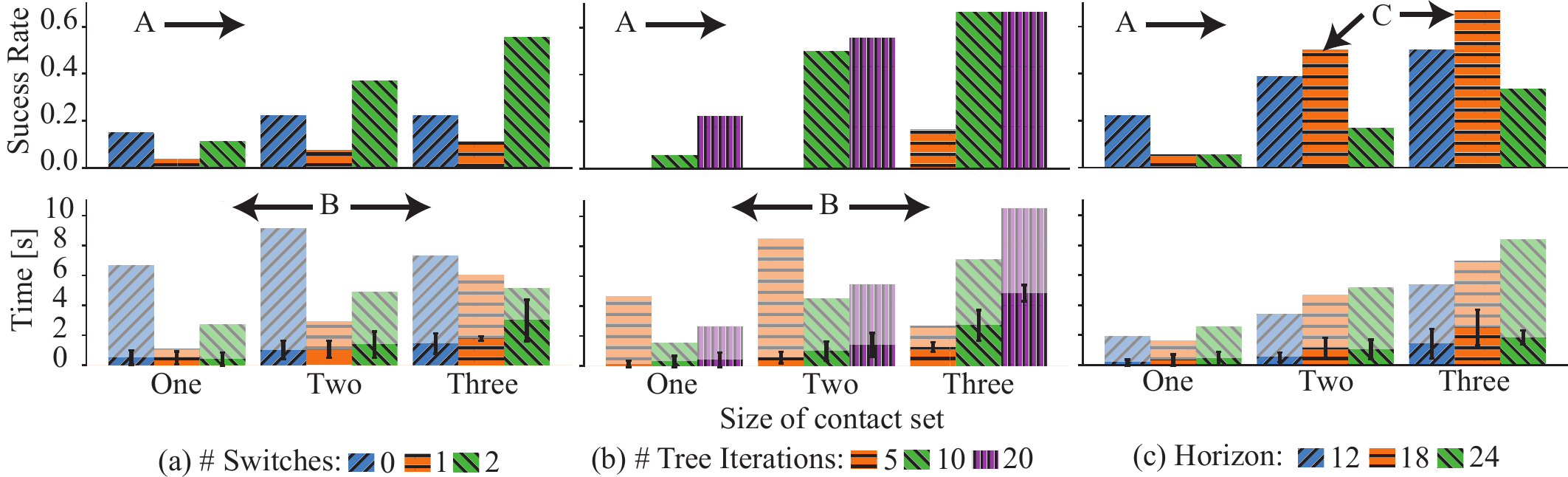}
\vspace{-0.25cm}
\caption{Ablation studies for planar pivoting following the convention in Fig.~\ref{fig:pp_ablation}. }
\label{fig:pl_ablation}
\vspace{-0.4cm}
\end{figure*}

\myparagraph{Planar Pushing}
We compute trajectories from eight initial conditions for enabled contact sets of size one and three (Fig.~\ref{fig:pp_traj}). The goal is ($x, y, \theta$) $=$ $(0, 0, 0)$. Trajectories are considered successful if the final errors in $x$, $y$, and $\theta$ are less than \SI{5}{\centi\meter}, \SI{5}{\centi\meter}, and \SI{5}{\degree}, respectively. We set the maximum number of hybrid switches ($N_\text{s}$) to 1, the maximum iterations during tree generation ($N_\text{i})$ to 10, and the planning horizon ($N$) to 24. Moreover, we use a time-step ($\Delta t$) of \SI{0.5}{\second}, a coefficient of friction ($\mu$) of 0.3 at both frictional contacts, and allow a maximum normal force ($N_\text{max}$) of \SI{0.5}{\newton}.   

With only the left contact enabled (Fig.~\ref{fig:pp_traj}a), as expected, the algorithm finds solutions for initial conditions that are to the left of the goal. Note that this corresponds to pure input-constrained DDP. With three contacts enabled (Fig.~\ref{fig:pp_traj}b), the algorithm finds trajectories to the goal from all initial conditions. The algorithm usually only needs to select the best contact; however, it needs a hybrid switch for one trajectory (solid green). The mean planning time is \SI{0.40}{\second} and \SI{0.70}{\second} for one and three enabled contacts, respectively. 

\myparagraph{Planar Pivoting}
We compute trajectories for enabled contact sets of size one, two and three from two initial conditions (Fig.~\ref{fig:pl_traj}). The goal is $\theta=$\SI{10}{\degree} and $\dot{\theta}=$\SI{0}{\degree\per\second}. Successful trajectories have final errors in $\theta$ and $\dot{\theta}$ that are less than \SI{10}{\degree} and \SI{10}{\degree\per\second}, respectively. The object's mass is \SI{0.1}{\kilo\gram} and its density is uniform. We use $N_\text{s} =2$, $N_\text{i} = 10$, $N=16$, $\Delta t=$ \SI{0.05}{\second}, $\mu = 0.5$, and $N_\text{max}=$\SI{10}{\newton}.    

For pivoting, we observe that the ability to reason about contact switches is important. For example, we cannot pivot the object from \SIrange{80}{10}{\degree} with only a single contact enabled using pure input-constrained DDP (Fig.~ \ref{fig:pl_traj}a). Moreover, the planner finds different mode sequences with more than one enabled contacts (Fig.~ \ref{fig:pl_traj}b). Finally, the mean planning time is 0.67, 3.12, and \SI{7.30}{\second} for the trajectories  with one, two, and three enabled contacts, respectively.

\subsection{Ablation Studies}
\label{sec:ablation}

We also conduct one-dimensional ablation studies that explore how the hyper-parameters of our algorithm affect success rate (defined above) and planning time.

\myparagraph{Planar Pushing}
In Fig.~\ref{fig:pp_ablation}, we depict the effect of the number enabled contacts and one other ablation parameter: (a) number of hybrid switches, (b) number of DDP iterations during mode selection (tree generation), and (c) the horizon of the trajectory. When not varied, these parameters are fixed to $N_\text{s} = 1$, $N_\text{i} = 5$, and $N = 24$. For each parameter, we consider all active contact combinations and plan trajectories from 27 initial conditions for each contact combination. 

We find that across all parameters, success rate increases with the number of enabled contacts (A, Fig.~\ref{fig:pp_ablation}). This is intuitive as allowing for more contact normal directions increases controllability. This success, however, comes with an increased planning time (B, Fig.~\ref{fig:pp_ablation}), though planning time is most affected by the number of hybrid switches (Fig.~\ref{fig:pp_ablation}a). We also find that success rate is insensitive to the choice of $N_i$ (Fig.~\ref{fig:pp_ablation}b), and is robust to all hyper-parameter changes with three or four enabled contacts (C, Fig.~\ref{fig:pp_ablation}). Finally, we can achieve a success rate of 100\% with a planning time of $\sim$\SI{1}{\second} for a number of different hyper-parameter combinations.

\myparagraph{Planar Pivoting}
We present the effects of varying the same hyper-parameters as above for pivoting in Fig.~\ref{fig:pl_ablation}. When not varied, these parameters are fixed to $N_\text{s} = 2$, $N_\text{i} = 10$, and $N = 18$. For each parameter combination, we consider all contact combinations and plan trajectories from two initial conditions for object with aspect ratios of 0.5, 1.0, and 1.5. 

Similar to the pushing primitive, we find that success rate increases with the number of enabled contacts (A, Fig.~\ref{fig:pl_ablation}); however, we are only able to reach a maximum success rate of 0.6-0.7. Interestingly, there is not a corresponding increase in planning time (B, Fig.~\ref{fig:pl_ablation}), though the overall planning time is higher than for planar pushing due to the more complex dynamics of pivoting. Our results suggest that the planner is also more sensitive to the choice of hyper-parameters; for example, planning over an 18-step horizon outperforms planning over 12 and 24-step horizons (C, Fig.~\ref{fig:pl_ablation}). 

\section{Experimental Results}
\label{sec:exp_results}

We evaluate our approach on a real planar pushing system. 

\begin{figure}[t]
    \centering
    \includegraphics[width=0.9\columnwidth]{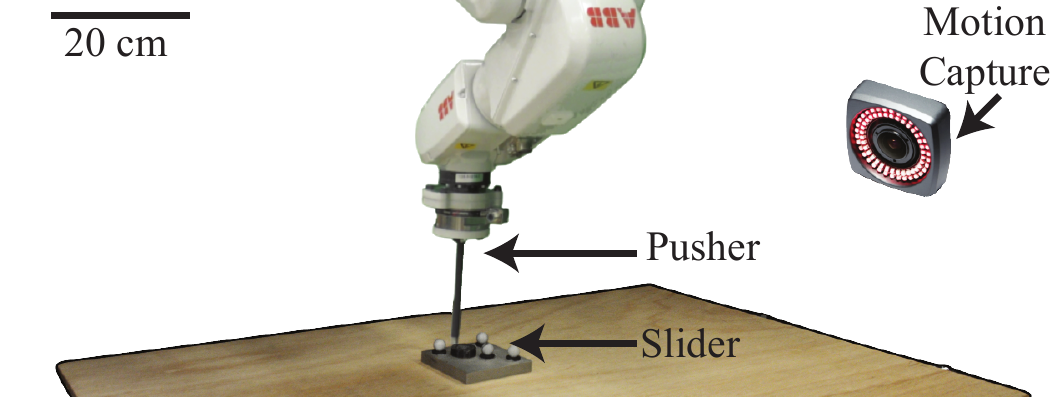}
    \vspace{-0.35cm}
    \caption{Robot arena for the planar pushing experiments.}
    \label{fig:exp_setup}
\end{figure}

\myparagraph{Experimental Set-up}
We use an industrial robotic manipulator (ABB IRB 120, Fig.~\ref{fig:exp_setup}). The object rests on a flat plywood surface and is moved by a metallic rod attached to the robot. The feedback controller \eqref{eq:control_update} runs at $\sim$\SI{250}{\hertz}, and the object pose is tracked using a motion capture system (Vicon, Bonita) at \SI{300}{\hertz}. The object's physical properties are described in Sec.~\ref{sec:rep_traj}, and it has a length of \SI{0.09}{\meter}.

We convert the inputs of our model (applied force and time-step length) into position commands for the robot manipulator by integrating the following kinematic relation, 
\begin{align}
\label{eq:force_to_pos}
\*{x}^p_{k+1} = \*{x}^p_k + \Delta t_k \*{R} \*{J}^i \*{L} (\*{J}^i)^T \*{f}^i_k.
\end{align}
Here $\*{x}^p$ is the Cartesian position of the contact in the world frame, and $\*{R}$, $\*{J}^i$, and $\*{L}$ are defined in Sec.~\ref{sec:pushing}.

\begin{figure}[t]
\centering
\includegraphics[width=0.95\columnwidth]{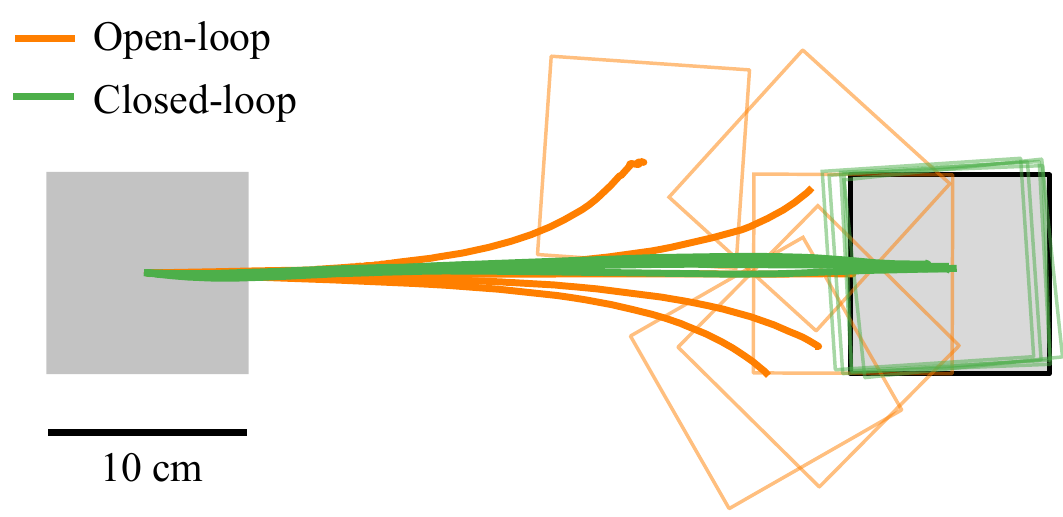}
\vspace{-0.55cm}
\caption{Open-loop (orange, n=5) and closed-loop (green, n=5) straight-line pushes. The light-gray box is the initial condition, and the black-outlined gray box is the goal. The controller significantly reduces error.}
\label{fig:control_results}
\vspace{-0.5cm}
\end{figure}

\myparagraph{Straight Line Pushing}
We evaluate the performance of our controller on straight-line pushing (Fig.~\ref{fig:control_results}). We execute five open-loop and five closed-loop  \SI{40}{\centi\meter} pushes. The open-loop standard deviation (s.d.) in error is $x=$\SI{3.5}{\centi\meter}, $y=$ \SI{4.2}{\centi\meter}, and $\theta=\SI{85}{\degree}$. The controller significantly reduces the closed-loop s.d. in error to $x=$\SI{0.5}{\centi\meter}, $y=$\SI{0.1}{\centi\meter}, and $\theta=$\SI{1.25}{\degree}.

\begin{figure}[t]
\centering
\includegraphics[width=0.9\columnwidth]{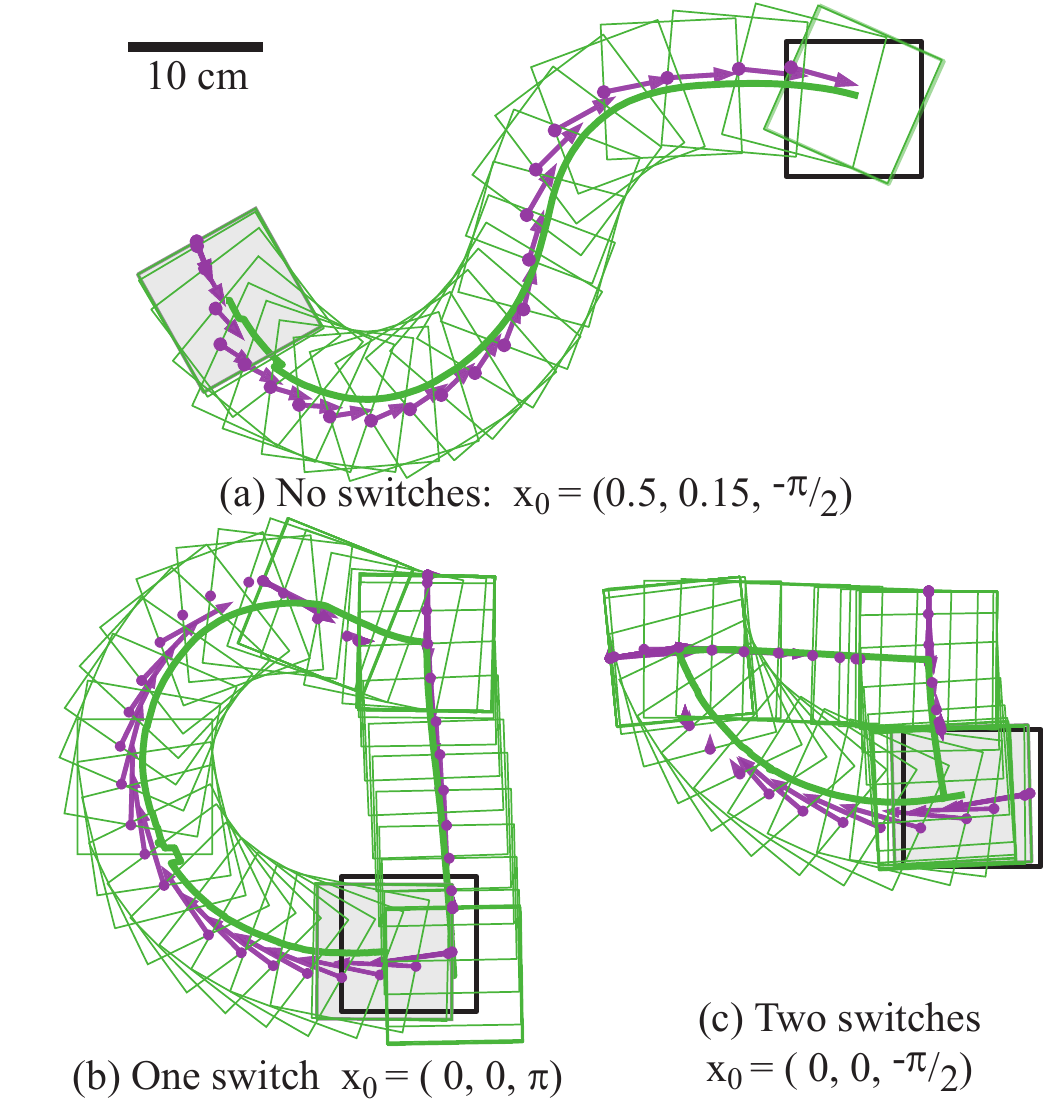}
\caption{Closed-loop pushes with (a) no contact switches, (b) one contact switch, and (c) two contact switches. The object pose and Cartesian trajectory is shown in green. Nominal contact locations and applied forces are shown with purple circles and arrows, respectively. The light-gray box is the initial condition, and the black-outlined box is the goal.}
\label{fig:exp_results}
\vspace{-0.35cm}
\end{figure}

\myparagraph{Hybrid Pushing}
We also validate our framework for three pushes starting from more challenging initial conditions, with zero, one, and two contact switches (Fig.~\ref{fig:exp_results}). Our planner finds pushing trajectories that reach the goal and are effectively stabilized by the controller. However, slipping between the pusher and the object results in slightly higher final pose error than in the straight-line pushing scenario. 

\section{Discussion}
\label{sec:discussion}
 We summarize the major findings of this work (Sec.~\ref{sec:conc}), discuss some important limitations (Sec.~\ref{sec:limits}), and propose directions for future work (Sec.~\ref{sec:future_work}).

\subsection{Conclusions}
\label{sec:conc}
We develop a hybrid DDP algorithm for dynamical systems with frictional contact and discontinuous switches. Our approach reasons over a finite horizon, supports a small number of contact switches, and generates a linear stabilizing controller. Our approach can quickly generate closed-loop trajectories that drive most initial conditions to the goal for the two planar manipulation primitives considered. Finally, we demonstrate our controller's ability to track planned trajectories on a real pushing system.

\subsection{Limitations}
\label{sec:limits}
Though we can drive any initial condition to the origin for planar pushing, this is not the case for planar pivoting. We believe this is due to poor initialization; while static equilibrium is a fixed-point for quasi-static planar pushing, it is not the same for dynamic planar pivoting. Furthermore, though we can accurately track straight-line pushes on a real system, tracking errors are larger for more complex trajectories. This is likely a result of slipping between the pusher and object, and a lower-level controller that enforces sticking (e.g.,  \cite{hogan2020tactile}) would complement our approach. Finally, the computational cost of our algorithm increases combinatorially with the number of allowed switches; however, we show that a small number of switches is sufficient for executing for planar pushing and planar pivoting.

\subsection{Future Work}
\label{sec:future_work}
One extension is to apply our approach on a wider range of primitives, including pulling, prehensile pushing, rolling, tilting, etc. This will require both identifying appropriate mechanics models and adapting the hybrid DDP framework. In particular, we would like to improve both the initialization and pruning procedure for our algorithm and to reduce its dependence on user defined hyper-parameters. We would also like to explore more sophisticated controllers, as detailed by Hogan and Rodriguez \cite{hogan2016wafr}, to reason about contact-sliding relative to the object.

\section*{Acknowledgements}

This research was supported by an appointment to the Intelligence Community Postdoctoral Research Fellowship Program at the Massachusetts Institute of Technology, administered by Oak Ridge Institute for Science and Education through an interagency agreement between the U.S. Department of Energy and the Office of the Director of National Intelligence.

\bibliographystyle{IEEEtran}
\bibliography{bibliography}

\end{document}